\documentclass{article}
\usepackage{adjustbox}
\usepackage{ijcai23}
\usepackage{dsfont}
\usepackage{colortbl}
\usepackage{array}
\usepackage{color}
\usepackage{times}
\usepackage[dvipsnames]{xcolor}
\usepackage{soul}
\usepackage{url}
\usepackage{subfigure}
\usepackage[hidelinks]{hyperref}
\usepackage[utf8]{inputenc}
\usepackage[small]{caption}
\usepackage{graphicx}
\usepackage{amsmath}
\usepackage{amsthm}
\usepackage{amssymb}

\usepackage{booktabs}
\usepackage{algorithm}
\usepackage{algorithmic}
\usepackage{multirow}
\usepackage[switch]{lineno}
\usepackage{bbding}

\title{FedTAD: Topology-aware Data-free Knowledge Distillation for Subgraph Federated Learning}

\author{
Yinlin Zhu$^1$\and
Xunkai Li$^2$\and
Zhengyu Wu$^{2}$\and
Di Wu$^{1}$\footnotemark[2]
\and
Miao Hu$^1$\and
Rong-Hua Li$^2$
\\
\affiliations
$^1$Sun Yat-sen University, Guangzhou, China\\
$^2$Beijing Institute of Technology, Beijing, China\\
\emails
zhuylin27@mail2.sysu.edu.cn,
cs.xunkai.li@gmail.com,
Jeremywzy96@outlook.com,\\
\{wudi27, humiao\}@mail.sysu.edu.cn,
lironghuabit@126.com
}

\begin{document}

\maketitle
\footnotetext[2]{Corresponding author.}
\begin{abstract}
    Subgraph federated learning (subgraph-FL) is a new distributed paradigm that facilitates the collaborative training of graph neural networks (GNNs) by multi-client subgraphs. Unfortunately, a significant challenge of subgraph-FL arises from subgraph heterogeneity, which stems from node and topology variation, causing the impaired performance of the global GNN. Despite various studies, they have not yet thoroughly investigated the impact mechanism of subgraph heterogeneity. To this end, we decouple node and topology variation, revealing that they correspond to differences in label distribution and structure homophily. Remarkably, these variations lead to significant differences in the class-wise knowledge reliability of multiple local GNNs, misguiding the model aggregation with varying degrees. Building on this insight, we propose topology-aware data-free knowledge distillation technology (FedTAD), enhancing reliable knowledge transfer from the local model to the global model. Extensive experiments on six public datasets consistently demonstrate the superiority of FedTAD over state-of-the-art baselines.
\end{abstract}

\section{Introduction}

%------------------------------------------------------------------------------------

    Graph neural networks (GNNs) have emerged as a promising machine learning paradigm on structured data, exhibiting remarkable performance across diverse AI applications, such as social recommendation ~\cite{guo2020deep} and financial analysis~\cite{yang2021financial}. 
    However, most existing GNNs presuppose centralized data storage, which grants a single user or institution access to the entire graph for training.
    This vulnerable assumption does not apply to many real-world scenarios, as a domain-specific graph may consist of multiple subgraphs, each constrained to be locally accessed due to competition and privacy concerns. 
    For instance, each financial institution constructs its transaction network without sharing privacy-sensitive data with others~\cite{fu2022federated}. 
    Consequently, training powerful GNNs using collective intelligence becomes challenging in such distributed scenarios.

%------------------------------------------------------------------------------------

    To this end, federated graph learning (FGL), the concept of integrating federated learning (FL) with graphs is proposed.
    Especially, the instance of FGL on a semi-supervised node classification paradigm is known as subgraph-FL. 
    In each training round of this federated setting, each client utilizes its privately stored subgraph to train the local model independently.
    Then, these local models are uploaded to a trusted central server and aggregated into a global model, which will be broadcast to local clients for the next training round.
    In the aforementioned multi-client collaborative training process, FedAvg~\cite{mcmahan2017communication} serves as a simple yet effective method, accomplishing global model aggregation by quantifying the sizes of data samples across multiple clients.

%------------------------------------------------------------------------------------
    
    Despite extensive study of subgraph-FL, inherent differences in data collection methods often lead to multi-client data variation~\cite{xie2021federated}, leading to a decline in the model performance, referred to as the \textit{subgraph heterogeneity}~\cite{baek2023personalized}. 
    Unlike data heterogeneity in conventional FL, subgraph heterogeneity arises from both nodes and topology, as shown in Fig.~\ref{empirical_result}(a). As a result, the conventional federated optimization strategies fail at addressing subgraph heterogeneity (e.g., FedProx~\cite{li2020federated} for Non-iid label distributions) due to the emission of topology.
    To this end, Fed-PUB~\cite{baek2023personalized} measures subgraph similarity for personalized aggregation. 
    FedGTA~\cite{xkLi_FedGTA_VLDB_2024} introduce graph mixed moments for topology-aware aggregation.
    FedSage+\cite{zhang2021subgraph} and FedGNN\cite{wu2021fedgnn} aspire to reconstruct potential missing edges among clients, representing an implicit resolution to subgraph heterogeneity by aligning local optimization objectives at the data level.
    Despite the considerable efforts of subgraph heterogeneity, they still have the following limitations. 

%------------------------------------------------------------------------------------

\textbf{L1: Lack of in-depth exploration.}
    The impact mechanisms of subgraph heterogeneity from the independent roles of node and topology variation are ambiguous.
    \textbf{Solution:}
    we empirically investigate and decouple node and topology variation among clients, revealing that they correspond to the difference in label distributions and class-wise homophily (i.e., for a specific class of nodes, the preference of them and their neighbors with similar attributes). These variations lead to different class-wise knowledge reliability. Specifically, for a specific label, the local model trained with a local subgraph that has numerous nodes with this label and exhibits strong class-wise homophily has a reliable prediction.
    
\textbf{L2: Sub-optimal model aggregation in subgraph-FL.}
    Due to the local training process encoding patterns of local subgraph into parameters of the local model, class-wise knowledge with different reliability is implicitly contained in these parameters. As a result, the server-side model aggregation is inevitably misled by unreliable class-wise knowledge.
    \textbf{Solution:} 
    we propose FedTAD, as a topology-aware data-free knowledge distillation strategy, which fully considers the differences in the reliability of class-wise knowledge among local GNNs to improve subgraph-FL. 
    Specifically, on the \underline{\textit{Client side}}, FedTAD first utilizes topology-aware node embeddings to measure the reliability of class-wise knowledge. 
    Then, on the \underline{\textit{Server side}}, FedTAD employs a generator to model the input space and generates a pseudo graph for transferring reliable knowledge from the multi-client local model to the global model. 
    Remarkably, FedTAD can be viewed as a hot-plugging strategy for any FL optimization strategy, aiming to correct the global model misled by unreliable knowledge during multi-client model aggregation.

    \textbf{Our contributions.}
    (1) \underline{\textit{New Observation.}} 
    To the best of our knowledge, this paper is the first to investigate the subgraph heterogeneity by decoupling multi-client node and topology variation, providing valuable empirical analysis.
    (2) \underline{\textit{New Method.}} 
    We propose FedTAD, which can serve as a hot-plugging post-processor for existing federated optimization strategies, achieving reliable knowledge transfer and alleviating the negative impact of unreliable class-wise knowledge caused by subgraph heterogeneity.
    (3) \underline{\textit{SOTA Performance.}} Extensive experiments on six datasets demonstrate that FedTAD outperforms existing baselines (up to 5.3\% higher), and improves various state-of-the-art FGL optimization strategies (up to 4.9\% performance boost).

\section{Preliminaries}

\paragraph{Graph Neural Networks.}    Consider an undirected graph $G=(\mathcal{V},\mathcal{E})$ with $\mathcal{V}$ as the node-set and $\mathcal{E}$ as the edge-set. Each node $v_i \in \mathcal{V}$ has a $F$-dimensional attribute vector $\boldsymbol{x}_i$ and a label $y_i$. The adjacency matrix (including self-loops) is ${\hat{\mathbf{A}}} \in \mathbb{R}^{N\times N}$. $\mathcal{N}(v_i)$ represents the neighbor nodes set of node $v_i$.
    Building upon this, most GNNs can be subsumed into the deep message-passing framework~\cite{gilmer2017neural}, which obtains a representation of the current node by recursively aggregating and transforming the representations of its neighbors. 
    Specifically, the representation of node $v_i$ at the $l$-th layer is denoted as $h_i^{l+1}$ and is computed as follows:
    \begin{equation}
    \begin{aligned}
    \boldsymbol{h}_i^{l+1}=\text{UPD}(\boldsymbol{h}_i^{l}, \text{AGG}(\{\boldsymbol{h}_j^l: v_j\in\mathcal{N}(v_i)\})),
    \end{aligned}   
    \end{equation}
    where $\boldsymbol{h}_i^0=\boldsymbol{x}_i$, $\boldsymbol{h}_i^{l}$ is the representation of node $v_i$ in the $l$-th (previous) layer, $\text{AGG}(\cdot)$ aggregates the neighbor representations, and $\text{UPD}(\cdot,\cdot)$ updates the representation of the current node using its representation and the aggregated neighbor representation at the previous layer.

\paragraph{Subgraph Federated Learning.}
    In subgraph-FL, the $k$-th client has an undirected subgraph $G_k=(\mathcal{V}_k, \mathcal{E}_k)$ of an implicit global graph $G_g=(\mathcal{V}_g, \mathcal{E}_g)$ (i.e., $\mathcal{V}_k \subseteq \mathcal{V}_g, \mathcal{E}_k \subseteq \mathcal{E}_g$).
    Each node $v_i \in \mathcal{V}_k$ has an attribute vector $\boldsymbol{x}_k^i$ and a label $y_k^i$. Typically, the training process of the $t$-th communication round in subgraph-FL with FedAvg aggregation strategy can be described as follows:
    (i) \textit{Client Selection}: 
    Randomly select a set of clients $S_t$ based on a certain probability for this round of subgraph-FL.
    (ii) \textit{Local Update}: Each selected client (e.g., the $k$-th) downloads the parameters of the global GNN $\widetilde{\boldsymbol{w}}^t$ and performs local training on its private subgraph $G_k$ to minimize $\mathcal{L}(G_k)$, where $\mathcal{L}(G_k)$ is the task loss for semi-supervised node classification on subgraph $G_k$.
    (iii) \textit{Global Aggregation}: 
    The server collects the parameters of the local GNNs $\{w_k\}_{k\in S_t}$ to perform parameter aggregation concerning the number of training instances, i.e., $\widetilde{\boldsymbol{w}}^t = \sum_{k\in S_t} \frac{|\mathcal{V}_k|}{N} \boldsymbol{w}_k^t$, where $N$ denotes the total number of nodes for all selected clients.

\paragraph{Class-wise Homophily.}
Various GNN research \cite{zhu2020beyond,sun2023breaking} has discussed the graph homophily (i.e., for an entire graph, the preference of connected nodes with similar attributes). However, this paper focuses on a more fine-grained issue: class-wise graph homophily (i.e., for a specific class of nodes, the preference of them and their neighbors with similar attributes). We contend that local GNNs offer more reliable predictions for classes with higher class-level homophily, providing effective guidance for subgraph-FL. Details are presented in Section \ref{empirical_study_section}.

\section{Empirical Investigation}\label{empirical_study_section}

\begin{figure*}
\rmfamily
\centering
\subfigure[Subgraph Heterogeneity]{
\includegraphics[width=0.22\textwidth]{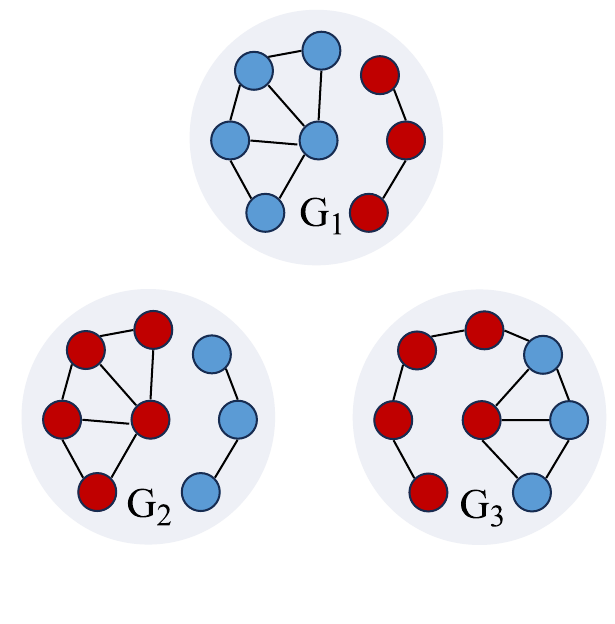} 
}
\subfigure[Data Simulation]{
\includegraphics[width=0.18\textwidth]{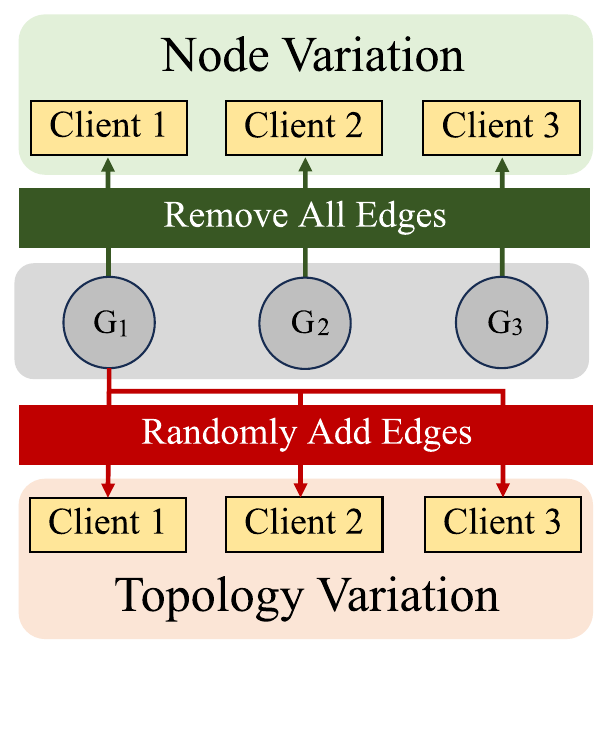} 
}\hspace{-2mm}
\subfigure[Label and Homophily Distributions]{
\includegraphics[width=0.27\textwidth]{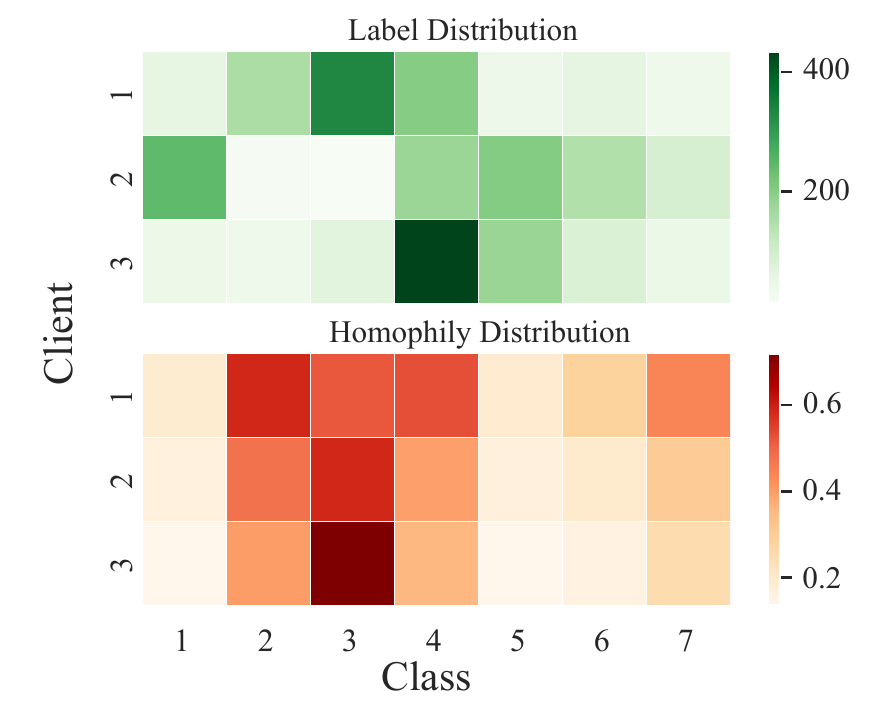} 
}\hspace{-2mm}
\subfigure[Performance Comparisons]{
\includegraphics[width=0.27\textwidth]{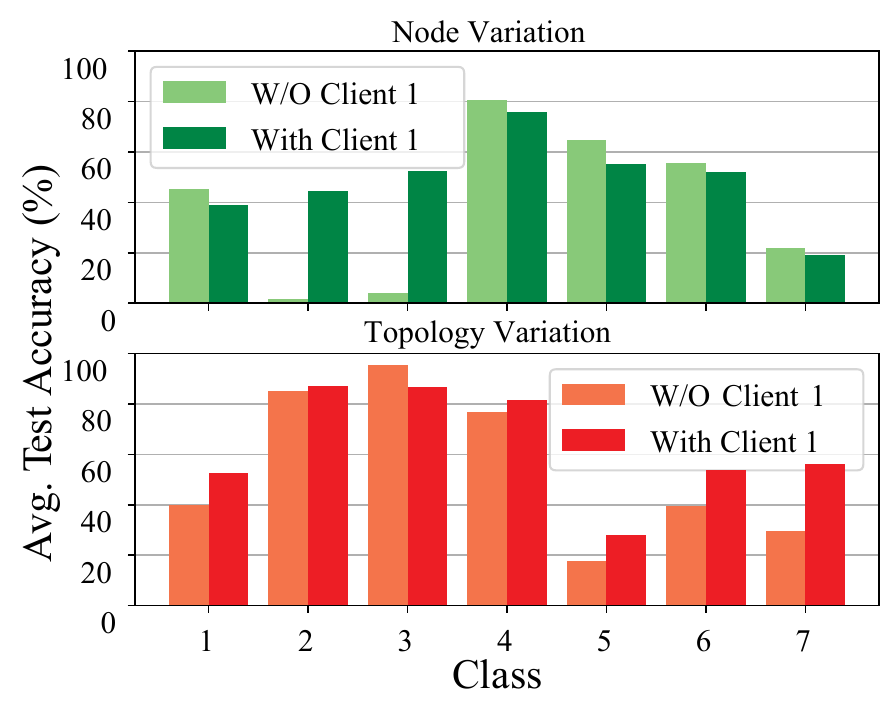} 
}\hspace{-2mm}
\DeclareGraphicsExtensions.
\vspace{-5pt}
\caption{(a) Illustration of subgraph heterogeneity. $G_1$ and $G_2$ exhibit different node label distributions (i.e., node variation); $G_2$ and $G_3$ have the same node label distributions but significant differences in topological properties (i.e., topology variation). (b) Two data simulation methods of our empirical study. (c) Upper: label distribution under the node variation scenario, the deeper color corresponds to a larger number of nodes; Lower: class-wise homophily distribution under the topology variation scenario, the deeper color corresponds to a stronger class-wise homophily. (d) Performance of global models with/without Client 1 participation under the node variation scenario (upper) and the topology variation scenario (lower).}
\label{empirical_result}
\vspace{-10pt}
\end{figure*}

    In this section, we delve into subgraph heterogeneity by separating node and topology entanglement. We first apply the average assignment-based Louvain algorithm~\cite{blondel2008fast} to split the Cora dataset \cite{yang2016revisiting} into three subgraphs, aligning with the strategy in FedSage+~\cite{zhang2021subgraph}. Before being distributed to clients, these subgraphs are modified to decouple the node and topology variation, which are detailed in Fig.~\ref{empirical_result}(b). 
    (i) \underline{\textit{Node variation.}} Removing edges to mitigate topology variation effects. 
    (ii) \underline{\textit{Topology variation.}} Clients share the nodes of subgraph 1, eliminating node variation, while the topology is perturbed by randomly adding edges.
    Then, we conduct subgraph-FL using GCN and FedAvg, comparing global GNN performance with/without Client 1 participation.

\paragraph{Node Variation.}
    Essentially, node variation can be regarded as the data heterogeneity in conventional FL, manifested as label Non-independent identical distribution (Non-iid). Node heterogeneity and the performance of two global models (with/without Client 1 participation) are depicted in Fig.~\ref{empirical_result}(c) and (d).
    Notably, client 1 predominantly collects nodes of classes 2 and 3, while other class nodes are scarce on Client 1.     
    Consequently, global model performance is notably improved for classes 2 and 3 with Client 1 participation, but it declines to some extent for other classes.
    Thus, we argue that the node variation leads to differences in the reliability of class-wise knowledge. 
    \textbf{Observation 1.}
    For a specific class, local GNN trained with a larger number of nodes corresponds to more reliable knowledge.

\paragraph{Topology Variation.}
    Building on the previous exploration, where node variation influences class-level knowledge reliability within local GNNs, we assume that topology variation may yield similar effects.
    Yet, a critical challenge is properly mapping the intricate topology into class-level distribution.
    Referring to a previous study~\cite{zhu2020beyond}, GNNs generally exhibit improved performance when dealing with graphs characterized by higher homophily. 
    Intuitively, we assume that GNNs offer more reliable class-wise knowledge when the majority of nodes within that class exhibit stronger local homophily.
    Consequently, we adapt the \textit{edge homophily ratio} \cite{zhu2020beyond} to quantify class-wise homophily.
    Specifically, consider a graph $G=(\mathcal{V}, \mathcal{E})$ with node label $y_i$ for each node $v_i \in \mathcal{V}$,
    we use $\boldsymbol{\varphi}(c)$ to denotes the class-wise homophily on class $c$, which is computed as:
\begin{equation}
\begin{aligned}
\label{class_wise_homo_eq}
\boldsymbol{\varphi}(c)\! = \!\frac{|\{(v_p,v_q)\!:\!(v_p,v_q)\in \mathcal{E} \wedge (y_p = c \wedge y_q = c)\}|}{|\{(v_p,v_q)\!:\!(v_p,v_q)\in \mathcal{E} \wedge (y_p = c \vee y_q = c)\}|}.
\end{aligned}
\end{equation}
    Fig.~\ref{empirical_result}(c) and (d) display the class-wise homophily distribution of three clients and the performance of two global models (with/without Client 1 participation).
    Notably, Client 1 exhibits weak class-wise homophily in class 3 but strong homophily in other classes. Global model performance for class 3 declines with Client 1 participation but improves for other classes. We assert that topology variation corresponds to differences in class-wise homophily and also leads to variations in the reliability of class-wise knowledge.
    \textbf{Observation 2.}
    For a specific class, local GNN trained with stronger class-wise homophily corresponds to more reliable knowledge.

\section{Methodology}

\begin{figure*}
  \centering
  \includegraphics[width=0.95\textwidth]{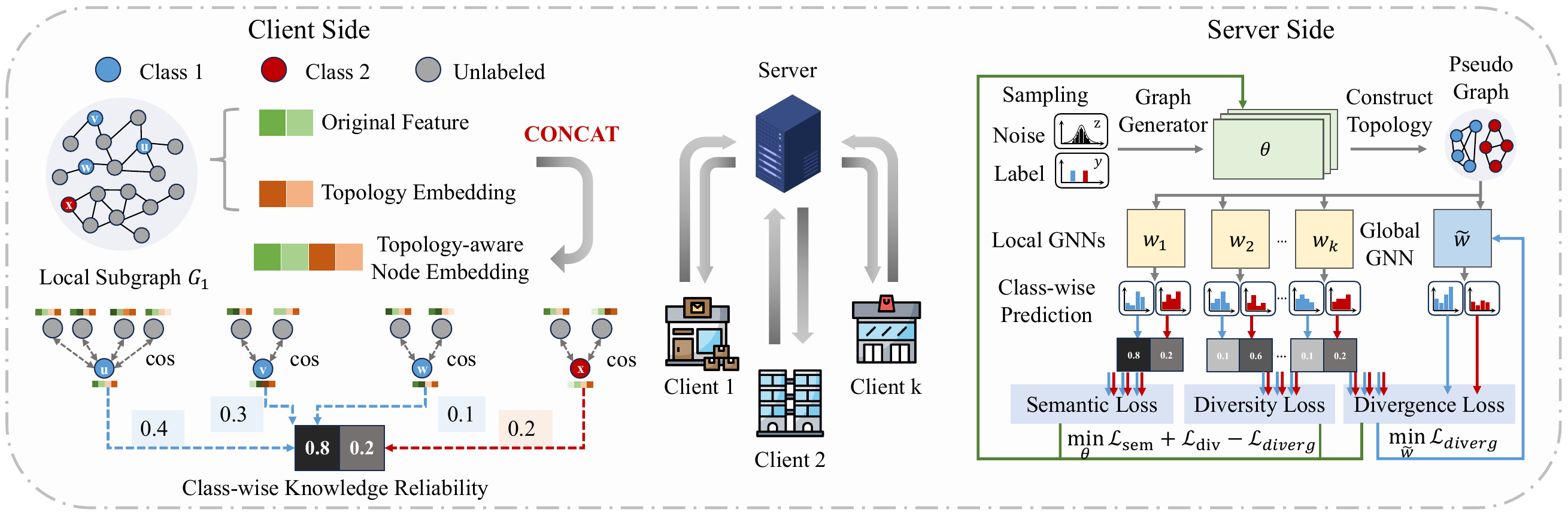}%
  \caption{An overview of our proposed FedTAD framework. On the client side, each client performs local initialization for measuring class-wise knowledge reliability. On the server side, the FedTAD can be regarded as a post-process of vanilla model aggregation, which enhances reliable class-wise knowledge transferring from local models to the global model.}
  \label{framework}
  \vspace{-10pt}
\end{figure*}

\subsection{Overview}
    The overview of our proposed FedTAD is depicted in Fig.\ref{framework}.
    On the \underline{\textit{Client side}}, each client computes topology-aware node embeddings to measure class-wise knowledge reliability, which is then uploaded to the server.
    On the \underline{\textit{Server side}}, guided by the class-wise knowledge reliability, FedTAD generates a pseudo graph for transferring reliable knowledge from multi-client local models to the global model.

\subsection{Class-wise Knowledge Reliability Measuring}
    To obtain reliable knowledge from clients, our key insight is that the reliability of local models for specific label predictions depends on their local subgraph (i) a large number of corresponding labeled nodes and (ii) strong class-wise graph homophily.
    However, the existence of unlabeled nodes hinders access to class-wise graph homophily. To this end, we introduce a topology-aware node embedding, which characterizes a node with its original feature and local topological structure. Afterward, we assess the class-wise knowledge reliability by quantifying the similarity in topology-aware node embeddings between labeled nodes and their neighbors.
\paragraph{Topology-aware Node Embedding.} We obtain the topology embedding of each labeled node $i$ based on the graph diffusion~\cite{tong2006fast}, which is defined as:
\begin{equation}
\begin{aligned}
\label{topo_emb_eq}
\boldsymbol{h}_i^{\text{TOPO}} = [\boldsymbol{T}_{ii}^1, \boldsymbol{T}_{ii}^2, ..., \boldsymbol{T}_{ii}^p] \in \mathbb{R}^{p},
\end{aligned}
\end{equation}
where $\boldsymbol{T}=\boldsymbol{A}\boldsymbol{D}^{-1}$ represents the random walk transition matrix, $\boldsymbol{D}$ represents the degree matrix (i.e., $\boldsymbol{d}_i = \sum_{j=1}^n \boldsymbol{a}_{ij}$). 
    By combining different walking distances $\{1,...,p\}$, the topology embedding captures rich structural information.
    Then, the topology-aware node embedding of $v_i$ can be obtained by concatenating the node features $\boldsymbol{x}_i$ and $\boldsymbol{h}_i^{\text{TOPO}}$:
\begin{equation}
\begin{aligned}
\label{hyb_emb_eq}
\boldsymbol{h}_i^{\text{HYB}} = \boldsymbol{x}_i|| \boldsymbol{h}_i^{\text{TOPO}}.
\end{aligned}
\end{equation}
\paragraph{Class-wise Knowledge Reliability.} Building upon the above conceptions, for a specific class $c$, the class-wise knowledge reliability is formally defined as:
\begin{equation}
\begin{aligned}
\label{class_wise_kr_eq}
\boldsymbol{\phi}_{c} = \sum_{v_i \in \mathcal{V}^{\text{labeled}}} \frac{\sum_{v_j \in \mathcal{N}(v_i)} \boldsymbol{s}(\boldsymbol{h}_i^{\text{HYB}}, \boldsymbol{h}_j^{\text{HYB}})}{|\mathcal{N}(v_i)|},
\end{aligned}
\end{equation}
where $\boldsymbol{s}(\cdot, \cdot)$ represents the cosine similarity. Intuitively, Eq.~(\ref{class_wise_kr_eq}) considers both the label and structure distribution, as the $\mathcal{V}^{\text{labeled}}$ term performs the summation of labeled nodes of the same class, and the ${\sum_{v_j \in \mathcal{N}(v_i)} \boldsymbol{s}(\boldsymbol{h}_i^{\text{HYB}}, \boldsymbol{h}_j^{\text{HYB}})}/{|\mathcal{N}(v_i)|}$ term measures the class-wise graph homophily within the 1-hop ego-graph of each labeled node.

\subsection{Topology-aware Knowledge Distillation}
    The proposed FedTAD servers as a post-processor following the vanilla model aggregators (e.g., FedAvg). 
    Its objective is to transfer reliable knowledge from multi-client local models to further improve the aggregated model $\boldsymbol{\widetilde{w}}$.
    Specifically, in each round $t$, the server receives the local model weights $\{\boldsymbol{w}_k\}_{k \in S_t}$ and the class-wise knowledge reliability $\{\boldsymbol{\phi}_k\}_{k \in S_t}$ uploaded from the selected clients $S_t$. 
    
\paragraph{Server-side Knowledge Generation.}
The server first uses a generator to generate pseudo node attributes $\boldsymbol{\hat{X}} \in \mathbb{R}^{B\times F}$,
\begin{equation}
\begin{aligned}
\label{x_eq}
\boldsymbol{\hat{X}} = P(\boldsymbol{z},\boldsymbol{y};\boldsymbol{\theta}),
\end{aligned}
\end{equation}
where $P$ denotes the generator, $\boldsymbol{\theta}$ is the parameter of generator, $\boldsymbol{z}$ is the standard Gaussian noise, i.e. $\boldsymbol{z}\sim\boldsymbol{\mathcal{N}}(\boldsymbol{0},\boldsymbol{1})$, $y$ denotes the label of $\boldsymbol{\hat{x}}$ sampled from uniform distribution, and $B$ denotes the number of generated nodes.

Subsequently, we construct the topology structure based on $\boldsymbol{\hat{X}}$ with the \textit{K-Nearest Neighbors} strategy. Specifically, the pseudo adjacency matrix $\boldsymbol{\hat{A}}$ is computed as follows:
\begin{equation}
\begin{aligned}
\label{knn_eq}
&\boldsymbol{H}\!=\! \sigma(\boldsymbol{\hat{X}}\boldsymbol{\hat{X}}^T\!),\;\!\boldsymbol{\hat{A}}[u,v]\!=\!\left\{
\begin{aligned}
&1,   \text{if }  v \in \text{TopK}(\boldsymbol{H}[u,:]); \\
&0,  \text{otherwise},
\end{aligned}
\right.
\end{aligned}
\end{equation}
where $\sigma(\cdot)$ represents the sigmoid function. The obtained pseudo graph is denoted as $\hat{G} = (\mathcal{\hat{V}}, \mathcal{\hat{E}})$ with the pseudo features $\boldsymbol{\hat{X}}$ and pseudo labels $\boldsymbol{\hat{Y}}$.

To ensure that generated samples closely resemble real ones, it is crucial to confine them within the same semantic space. In many data-free knowledge distillation studies in the vision domain, a strategy involves training the generator to be accurately predicted by teacher models \cite{chen2019data}. 
Building upon this approach, we further encourage the generator to learn reliable class-wise semantics from each local model by minimizing the following loss:
\begin{equation}
\begin{aligned}
\label{sem_obj_eq}
\mathcal{L}_{sem} = \sum_{k\in S_t}^K \sum_{c=1}^C\frac{\boldsymbol{\phi}_k^c}{\boldsymbol{\phi}^c} \sum_{v_i\in\hat{\mathcal{V}}_c}\mathcal{L}_\text{CE}(\sigma(f(v_i,\hat{G}; \boldsymbol{w}_k)), \boldsymbol{\hat{y}}_i),
\end{aligned}
\end{equation}
where $\boldsymbol{\phi}^c=\sum_{k\in S_t} \boldsymbol{\phi}_k^c$, $\hat{\mathcal{V}}_c$ denotes the pseudo node with label $c$, $\sigma(\cdot)$ is the softmax function, $f$ is the GNN classifier, and $\mathcal{L}_\text{CE}$ represents the cross-entropy loss. 
    Eq.~(\ref{sem_obj_eq}) dynamically adjusts the weights based on the global proportion of class-wise knowledge reliability, ensuring that the generated nodes can be precisely predicted by reliable clients.

    Furthermore, to mitigate the risk of mode collapse, the generator is encouraged to minimize diversity loss, facilitating the generation of diverse nodes, which is widely applied in generative models \cite{ding2018semi,mao2019mode}.
    \begin{equation}
    \begin{aligned}
    \label{mc_obj_eq}
    \mathcal{L}_{div}=\frac{\sum_{i,j\in \{1,..., B\}}}{B^2}\frac{\boldsymbol{\hat{x}}_i^T \boldsymbol{\hat{x}}_j}{|\!|\boldsymbol{\hat{x}}_i|\!|_2|\!|\boldsymbol{\hat{x}}_j|\!|_2}.
    \end{aligned}
    \end{equation}

\noindent Eq.~(\ref{mc_obj_eq}) can reduce the
similarity of features across pairs of pseudo nodes, aiming to
mitigate redundant features to prevent inefficiencies in subsequent
knowledge distillation.

\paragraph{Server-side Knowledge Distillation.} The pseudo graph is subsequently employed for knowledge distillation. 
    Owing to the global model being misled by unreliable knowledge, predictions of the global and local models exhibit notable differences in specific nodes. Consequently, for a specific local model, we incentivize the global model to minimize divergence on pseudo-nodes in reliable classes,
    \begin{equation}
    \begin{aligned}
    \label{div_obj_eq}
    \mathcal{L}_{diverg} = \sum_{k\in S_t}^K \sum_{c=1}^C\frac{\boldsymbol{\phi}_k^c}{\boldsymbol{\phi}^c}\sum_{v_i\in \mathcal{\hat{V}}_c}\text{KL}(f(v_i,\hat{G}; \boldsymbol{\widetilde{w}})||f(v_i,\hat{G}; \boldsymbol{w}_k)),
    \end{aligned}
    \end{equation}
where $\text{KL}(\cdot||\cdot)$ denotes the Kullback-Leibler divergence function. Conversely, the generator aims to generate pseudo nodes that induce divergence between the global and local model in reliable classes as much as possible. Therefore, the generation stage and the distillation stage compose an adversarial training process, which is formulated as follows:
\begin{equation}
\begin{aligned}
\label{tot_obj_eq}
\min\limits_{\boldsymbol{\widetilde{w}}}\max\limits_{\boldsymbol{\theta}}\mathbb{E}_{\boldsymbol{z} \sim \mathcal{\mathcal{N}}(\boldsymbol{0}, \boldsymbol{1})}[\mathcal{L}_{diverg} - \lambda_1\mathcal{L}_{sem} - \lambda_2\mathcal{L}_{div}],
\end{aligned}
\end{equation}
where $\lambda_1$ and $\lambda_2$ are trade-off parameters to control the contribution of different loss functions. The overall process is topology-aware since it is guided by class-wise knowledge reliability, considering class-wise graph homophily. The complete algorithm of FedTAD is presented in Algorithm~\ref{fedtad_algorithm}.

\begin{algorithm}
\caption{FedTAD Execution on Client and Server}
\label{fedtad_algorithm} 
\begin{algorithmic}[1]
\REQUIRE ~\\
Rounds, $T$; Local subgraphs, $\{G_k\}_{k\in S_t}$; Knowledge reliabilities, $\{\boldsymbol{\phi}_k\}_{k\in S_t}$; Server-side training iteration, $I$; generation iteration, $I_g$; distillation iteration, $I_d$;\\
\ENSURE ~\\
Global GNN weight, $\widetilde{\boldsymbol{w}}$;
\STATE \underline{\textit{Client-side Initialize}:}
\STATE Obtain topology-aware node embeddings via Eqs.~(\ref{topo_emb_eq}, \ref{hyb_emb_eq});
\vspace{-2.5ex}
\STATE Obtain the class-wise knowledge reliability via Eq.~(\ref{class_wise_kr_eq}).
\FOR{each communication round $t=1,...,T$}
\STATE \underline{\textit{Client-side Execute}:}
\STATE Update local GNN $\{\boldsymbol{w}_k\}_{k\in S_t}$ with $\widetilde{\boldsymbol{w}}$;
\STATE Train local GNN $\{\boldsymbol{w}_k\}_{k\in S_t}$ with $\{G_k\}_{k\in S_t}$;  
\STATE \underline{\textit{Server-side Execute}:}
\STATE Aggregate local GNN $\{\boldsymbol{w}_k\}_{k \in S_t} $ into global GNN $\widetilde{\boldsymbol{w}}$;
\FOR{each iteration $i=1,...,I$}
\STATE Generate pseudo node attributes $\boldsymbol{\hat{X}}$ via Eq.~(\ref{x_eq});
\STATE Obtain the pseudo adjacency $\boldsymbol{\hat{A}}$ via Eq.~(\ref{knn_eq});
\FOR{each inner iteration $g=1,...,I_g$}
\STATE Update the generator $\boldsymbol{\theta}$ via Eq.~(\ref{tot_obj_eq});  
\ENDFOR

\FOR{each inner iteration $d=1,...,I_d$}
\STATE Update the global model $\boldsymbol{\widetilde{w}}$ via Eq.~(\ref{tot_obj_eq});
\ENDFOR

\ENDFOR
\ENDFOR
\end{algorithmic}
\end{algorithm}

\section{Experiments}
In this section, we conduct experiments to verify the effectiveness of FedTAD. Specifically, we aim to answer the following questions: \textbf{Q1}: Compared with other state-of-the-art federated optimization strategies, can FedTAD achieve better performance? 
\textbf{Q2}: Where does the performance gain of FedTAD come from? \textbf{Q3}: Is FedTAD sensitive to the hyperparameters? \textbf{Q4}: Can FedTAD maintain robustness when partial clients participate? \textbf{Q5}: As a hot-plug method, how much performance improvement can FedTAD bring? \textbf{Q6}: What is the time complexity of FedTAD?

\subsection{Datasets and Simulation Method}
We perform experiments on six widely used public benchmark datasets in graph learning: three small-scale citation network datasets (Cora, CiteSeer, PubMed~\cite{yang2016revisiting}), two medium-scale co-author datasets (CS, Physics~\cite{shchur2018pitfalls}), and one large-scale OGB dataset (ogbn-arxiv~\cite{hu2020open}). More details can be found in Table \ref{dataset_detail}. To simulate the distributed subgraphs in the subgraph-FL, we employ the Louvain algorithm~\cite{blondel2008fast} to achieve graph partitioning, which is based on the optimization of modularity and widely used in subgraph-FL fields~\cite{zhang2021subgraph,he2021fedgraphnn}.

\subsection{Baselines and Experimental Settings}
\noindent{\textbf{Baselines}.} We compare the proposed FedTAD with five conventional FL optimization strategies (FedAvg~\cite{mcmahan2017communication}, FedProx~\cite{li2020federated}, SCAFFOLD~\cite{karimireddy2020scaffold}, MOON~\cite{li2021model}, FedDC~\cite{gao2022feddc}), two personalized subgraph-FL optimization strategies (Fed-PUB~\cite{baek2023personalized}, FedGTA~\cite{xkLi_FedGTA_VLDB_2024}), one personalized graph-FL optimization strategy (GCFL+~\cite{xie2021federated}), and one subgraph-FL framework (FedSage+~\cite{zhang2021subgraph}). 

\begin{table}
\rmfamily
\centering
\caption{\textrm{Statistics of the six public benchmark graph datasets.}}
\label{dataset_detail}
\resizebox{0.48\textwidth}{!}{
\begin{tabular}{cccccc} 
\toprule
\multicolumn{1}{c}{Dataset} & \multicolumn{1}{c}{\#Nodes} & \multicolumn{1}{c}{\#Features} & \multicolumn{1}{c}{\#Edges} & \multicolumn{1}{c}{\#Classes}  \\ 
\midrule
Cora                         & 2,708                     & 1,433                        & 5,429                     & 7                                \\
CiteSeer                     & 3,327                     & 3,703                        & 4,732                     & 6                                \\
PubMed                         & 19,717                     & 500                       & 44,338                     & 3                               \\
CS                          & 18,333                     & 6,805                        & 81,894                    & 15                                 \\
Physics                          & 34,493                    & 8,415                        & 247,692                    & 5                                 \\
ogbn-arxiv                         & 169,343                    & 128                         & 2,315,598                     & 40   \\
\bottomrule
\end{tabular}}
\end{table}

\noindent{\textbf{Hyperparameters}.} 
For each client and the central server, we employ a two-layer GCN as our backbone. 
The dimension of the hidden layer is set to 64 or 128.  
The local training epoch and round are set to 3 and 100, respectively.
The learning rate of GNN is set to 1e-2, the weight decay is set to 5e-4, and the dropout is set to 0.5. 
Based on this, we perform the hyperparameter search for FedTAD using the Optuna framework \cite{akiba2019optuna} on $\lambda_1$ and $\lambda_2$ within $\{10^{-1}, 10^{-2}, 10^{-3}\}$, and $I, I_g, I_d$ within $\{1, 3, 5, 10\}$. For each experiment, we report the mean and variance results of 3 standardized training.    

\begin{table*}
\caption{Performance comparison of test accuracy achieved by FedTAD and baseline models on six datasets. The best results are highlighted in \textbf{bold}, suboptimal results are marked with an \underline{underline}, and the third-best results are indicated by \colorbox{gray!20}{shading}.}
\centering
\label{baseline_comp}
\resizebox{1.0\textwidth}{!}{
\begin{tabular}{c|ccccccccccc}
\toprule
       Dataset $(\rightarrow)$      &            \multicolumn{3}{c}{Cora}            &  &           \multicolumn{3}{c}{CiteSeer}                    &  &            \multicolumn{3}{c}{PubMed}           \\
\cline{1-1} \cline{2-4}\cline{6-8}\cline{10-12}

\rule{0pt}{1.1em}Method $(\downarrow)$       & 5 Clients & 10 Clients   & 20 Clients &  & 5 Clients & 10 Clients & 20 Clients &  & 5 Clients & 10 Clients & 20 Clients \\ 
\midrule
FedAvg        & 80.6±0.3 & 73.6±0.4   & 56.0±0.3 &  & 71.5±0.3 & 68.9±0.2 & 66.3±0.4 &  & 85.6±0.3 & 82.9±0.0 & 80.6±0.3 \\

FedProx       & 80.9±0.2 & 73.3±0.3   & 56.3±0.4 &  & 71.3±0.2 & 69.1±0.3 & 65.9±0.4 &  & 85.3±0.8 & 82.7±0.1 & 80.5±0.4\\

SCAFFOLD      & 81.1±0.2 & 73.2±0.2   & 56.5±0.5 &  & 72.3±0.2 & 69.4±0.3 & 66.5±0.7 &  & 85.8±0.4 & 82.5±0.2 & 81.1±0.3  \\

MOON      & 81.5±0.2 & 73.3±0.4   & 56.6±0.3 &  & 71.8±0.2 & 68.4±0.3 & 66.7±0.2 &  & 86.1±0.3 & 82.5±0.3 & 80.4±0.2  \\

FedDC      & 81.3±0.4 & 73.4±0.3   & 56.3±0.4 &  & 71.5±0.4 & 69.3±0.2 & 67.1±0.6 &  & 85.4±0.2 & 82.2±0.5 & 80.8±0.4  \\
\midrule
GCFL+         & 81.7±0.3 & 73.8±0.5   & 56.6±0.2 &  & 72.0±0.3 & 69.8±0.3 & 67.5±0.2 &  & 85.8±0.3 & 83.3±0.3 & 81.3±0.4 \\

Fed-PUB      & 82.1±0.6 & 73.8±0.3   & 56.5±0.5 &  & 71.6±0.3 & 69.3±0.4 & 67.2±0.2 &  & 86.1±0.3 & 83.1±0.4 & 81.2±0.3 \\

FedSage+      & 82.7±0.5 &	73.9±0.2 &	58.1±0.7 &&		71.9±0.2	& 70.2±0.2 &	67.9±0.1 &&		86.1±0.4 &	\cellcolor{gray!20}83.8±0.3 &	81.6±0.7
 \\

FedGTA      & \cellcolor{gray!20}83.3±0.2 &	\cellcolor{gray!20}74.1±0.2 &	\cellcolor{gray!20}58.8±0.2 &&		\cellcolor{gray!20}72.4±0.2	& \cellcolor{gray!20}70.5±0.5 &	\cellcolor{gray!20}68.3±0.3 &&		\cellcolor{gray!20}86.3±0.3 &	83.7±0.3 &	\cellcolor{gray!20}82.0±0.4
 \\
\midrule
FedTAD (Ours) & \textbf{85.1±0.1} & \textbf{75.3±0.4}   & \textbf{61.3±0.3} &  & \textbf{73.5±0.3} & \textbf{71.7±0.4} & \textbf{70.2±0.3} &  & \textbf{87.9±0.1} & \textbf{84.4±0.4} & \textbf{83.5±0.2}\\ 
 
FedTAD (10\% Noise) & \underline{84.8±0.3} & \underline{75.1±0.5}   & \underline{60.9±0.3} &  & \underline{73.3±0.5} & \underline{71.4±0.2} & \underline{69.8±0.2} &  & \underline{87.7±0.4} & \underline{84.3±0.2} & \underline{83.2±0.3}\\ 
\midrule
Dataset $(\rightarrow)$            &            \multicolumn{3}{c}{CS}            &  &            \multicolumn{3}{c}{Physics}            &  &            \multicolumn{3}{c}{ogbn-arxiv}   \\ 
\cline{1-1} \cline{2-4}\cline{6-8}\cline{10-12}
\rule{0pt}{1.1em} Method $(\downarrow)$& 5 Clients & 10 Clients   & 20 Clients &  & 5 Clients & 10 Clients       & 20 Clients &  & 5 Clients & 10 Clients & 20 Clients\\ 
\midrule
FedAvg     & 90.4±0.3 &	85.9±0.8 &	83.9±0.3 && 94.7±0.3 &	91.6±0.0&	90.3±0.5 && 60.3±0.3&	58.4±0.2	& 55.7±0.3\\

FedProx      & 90.5±0.4 & 	86.1±0.7&	84.1±0.1&& 94.7±0.5&	91.6±0.2	& 90.3±0.0&& 60.2±0.2&	58.2±0.2&	55.5±0.1\\

SCAFFOLD     & 91.0±0.4 &	85.8±0.2 &	84.3±0.5 &&  94.9±0.3	& 91.4±0.3 &	90.6±0.0 && 60.4±0.1 &	58.7±0.5 &	55.9±0.8\\

MOON      & 91.2±0.3 & 85.7±0.4   & 84.1±0.7 &  & 95.1±0.4 & 91.3±0.2 & 91.1±0.3 &  & 60.4±0.2 & 58.6±0.7 & 55.5±0.4  \\

FedDC      & 91.4±0.5 & 86.1±0.3   & 84.5±0.5 &  & 94.6±0.7 & 91.7±0.4 & 91.5±0.4 &  & 60.9±0.2 & 58.5±0.4 & 56.1±0.5  \\
\midrule
GCFL+ &  91.9±0.2 &	86.3±0.1 &	84.5±0.4 &&		94.9±0.3 &	92.1±0.1 &	91.7±0.6 &&		61.5±0.2 &	58.1±0.1 &	55.9±0.4
  \\
Fed-PUB    & 90.8±0.4 &	86.7±0.4 &	84.7±0.4	&&	94.7±0.3 &	\cellcolor{gray!20}92.6±0.6 &	91.2±0.2 &&			
61.3±0.1 & 58.2±0.3 & 56.1±0.1
  \\
FedSage+ & 91.7±0.5 &	87.5±0.3 &	\cellcolor{gray!20}85.4±0.3	&&	94.8±0.4 &	92.2±0.4 &	91.8±0.4 &&    62.1±0.3       &   59.4±0.3        &   56.3±0.4    \\    
FedGTA  & \cellcolor{gray!20}92.0±0.2 &	\cellcolor{gray!20}88.7±0.0 &	85.2±0.3	&&	\cellcolor{gray!20}95.3±0.4 &	92.5±0.1 &	\cellcolor{gray!20}92.1±0.5 &&    \cellcolor{gray!20}62.4±0.3       &   \cellcolor{gray!20}60.5±0.2        &   \cellcolor{gray!20}57.3±0.1    \\
\midrule
FedTAD (Ours)        & \textbf{94.3±0.4}           &     \textbf{90.2±0.2}         &   \textbf{88.7±0.4}         &  &    \textbf{96.2±0.2}        & \textbf{94.1±0.2}                 &   \textbf{93.3±0.3}         &  &    \textbf{63.2±0.1}        &  \textbf{62.0±0.3}          &   \textbf{59.5±0.5} \\
FedTAD (10\% Noise) & \underline{94.1±0.3}           &     \underline{89.8±0.4}         &   \underline{88.3±0.2}         &  &    \underline{96.1±0.4}        & \underline{93.8±0.3}                 &   \underline{92.9±0.5}         &  &    \underline{62.9±0.3}        &  \underline{61.7±0.4}          &   \underline{59.1±0.3} \\
\bottomrule
\end{tabular}}
\end{table*}

\subsection{Experimental Analysis}

\begin{figure}[H]
\rmfamily
\centering
% \captionsetup[subfigure]{font=small}
% \captionsetup{skip=0.1pt}
\subfigure[Cora]{
\includegraphics[width=0.22\textwidth]{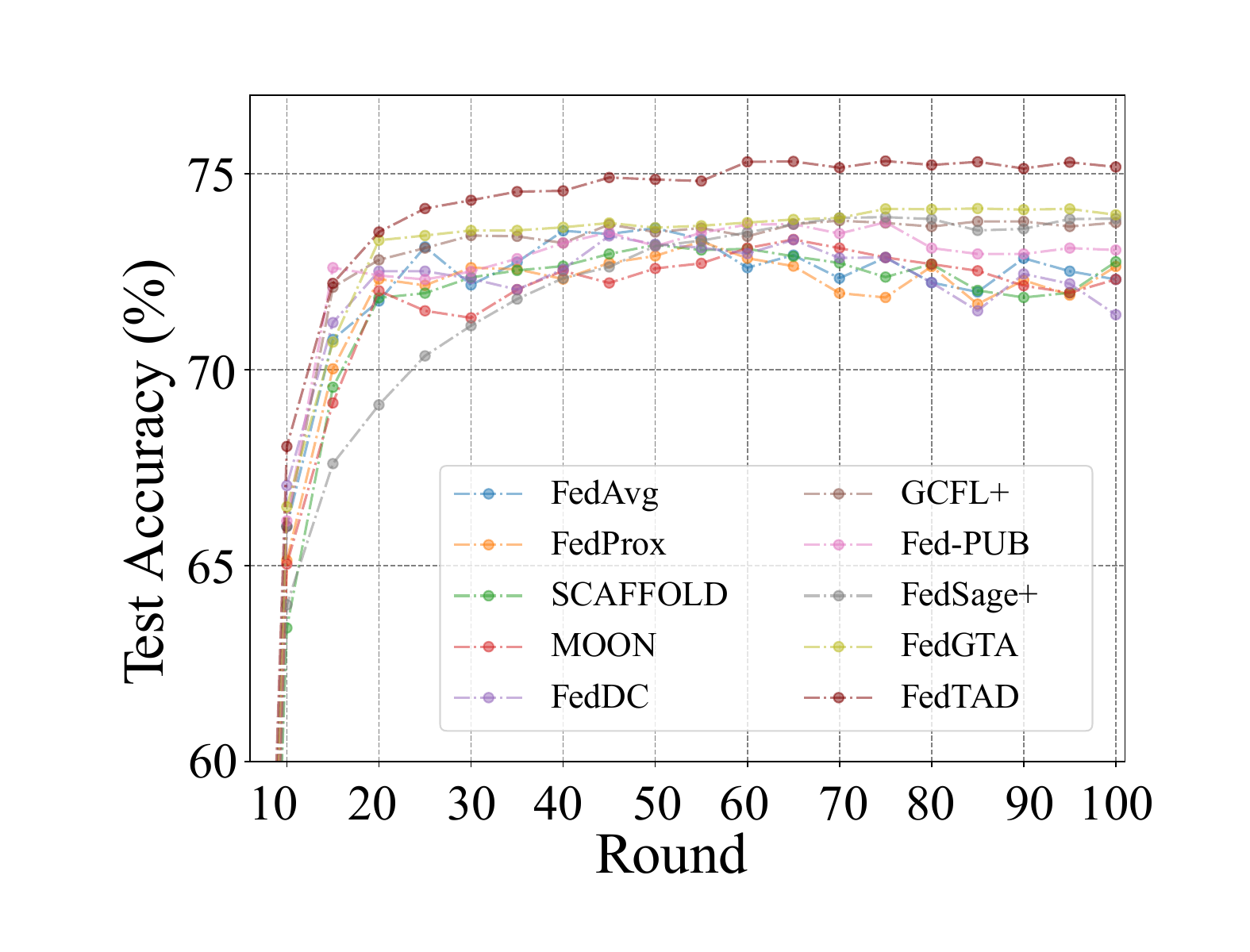} 
}
% \captionsetup{skip=0.1pt}
\subfigure[CiteSeer]{
\includegraphics[width=0.22\textwidth]{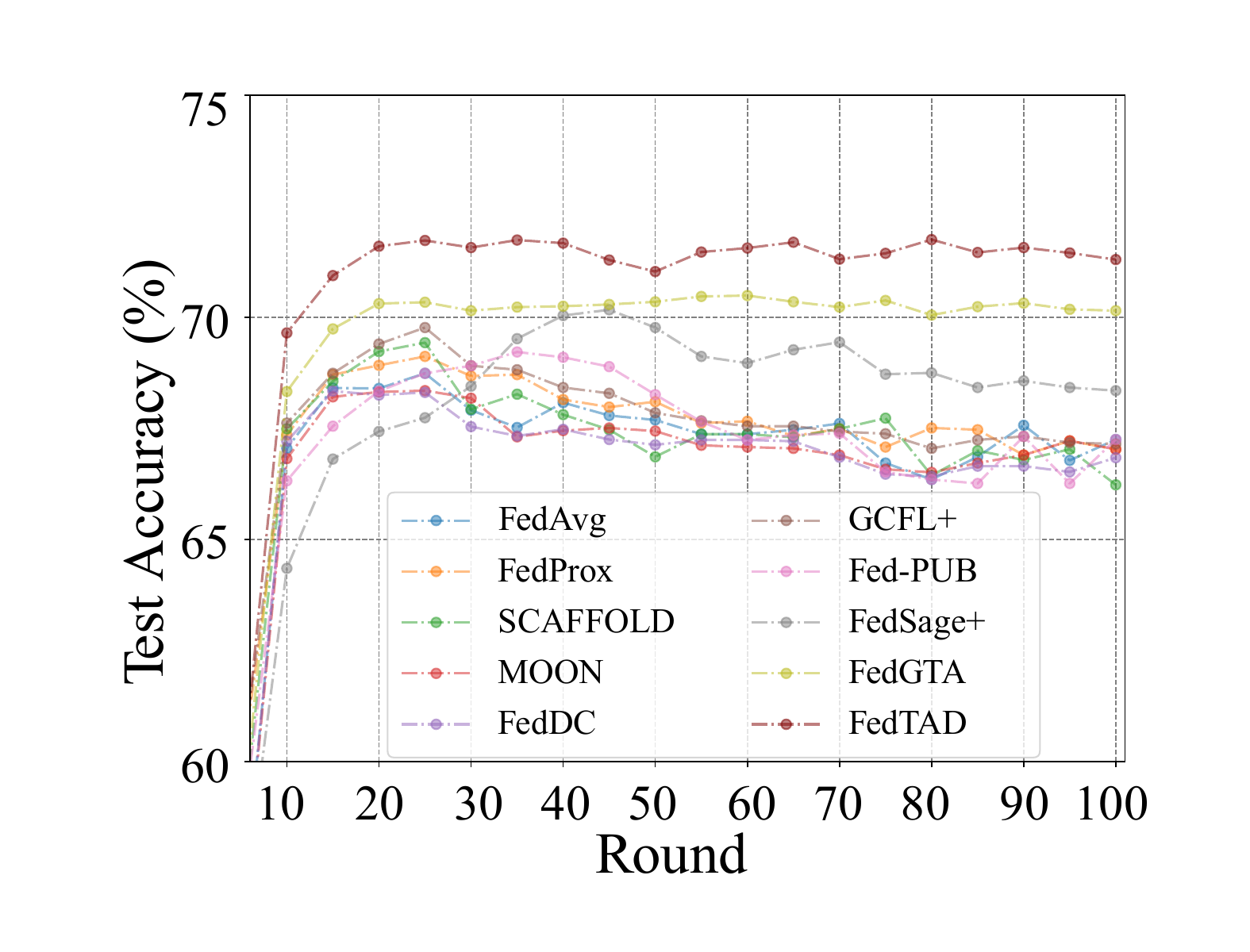} 
}
\\
\subfigure[PubMed]{
\includegraphics[width=0.22\textwidth]{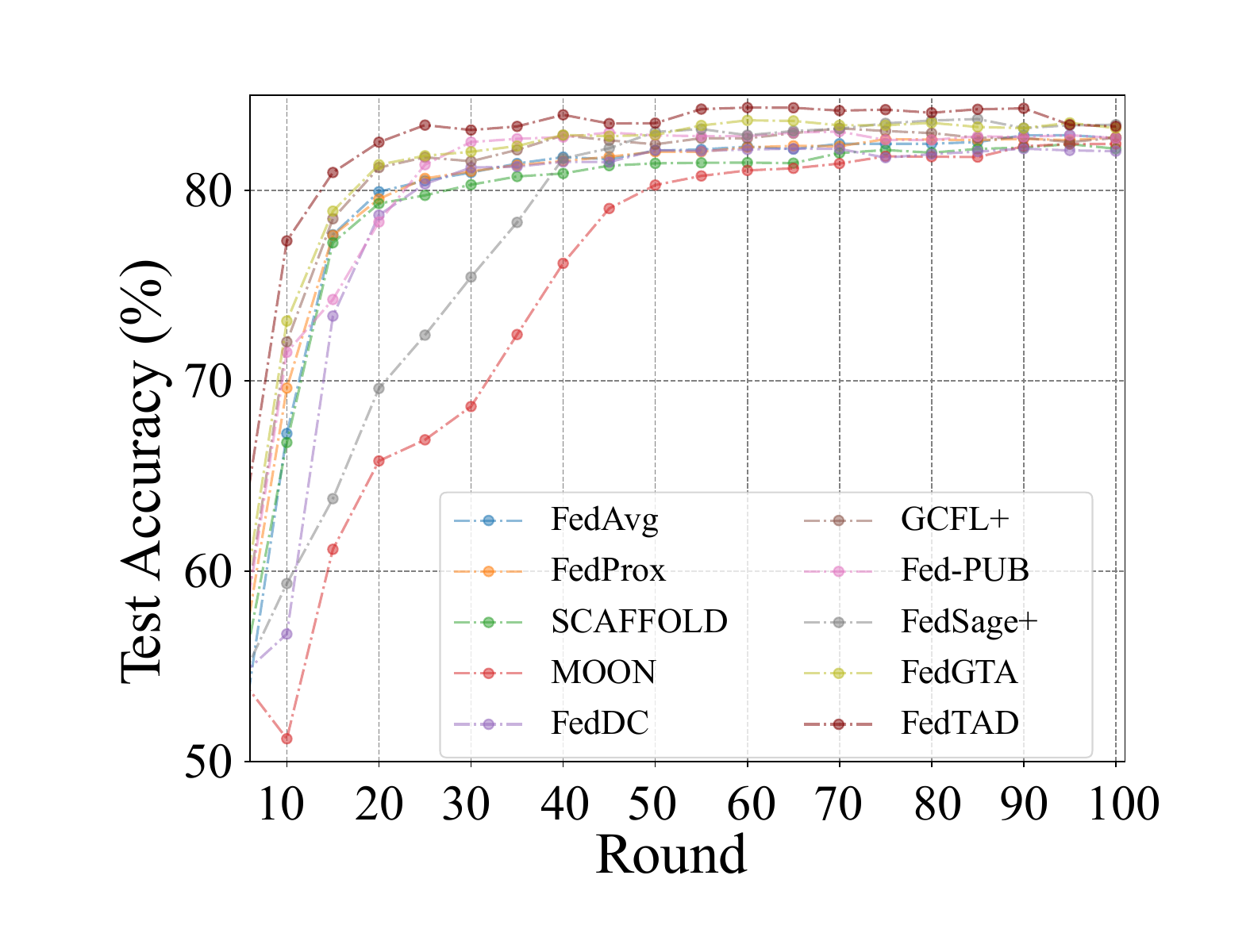} 
}
\captionsetup{skip=1.0pt}
\subfigure[ogbn-arxiv]{
\includegraphics[width=0.22\textwidth]{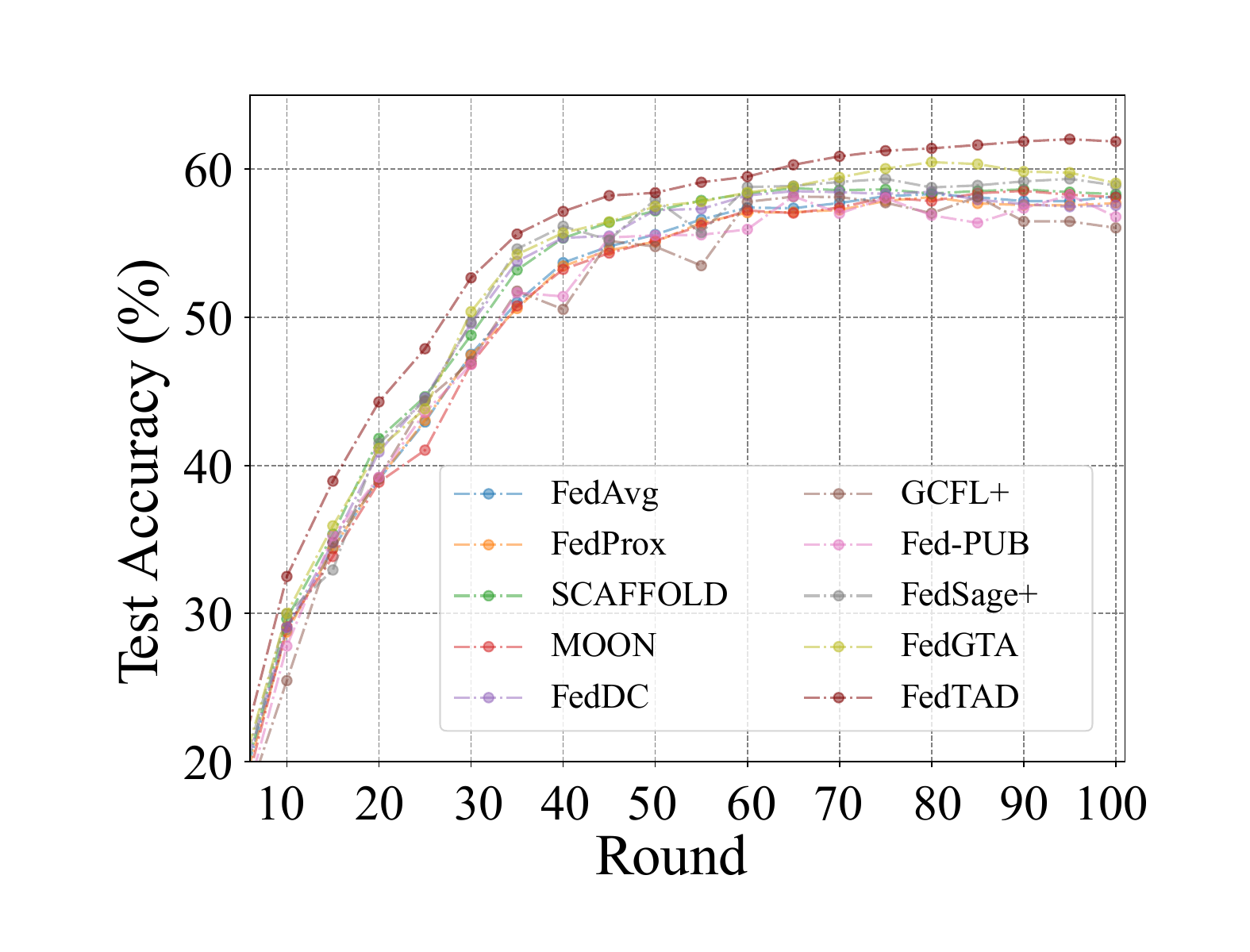} 
}
\DeclareGraphicsExtensions.
\caption{Convergence curves of our proposed FedTAD and baseline methods on four graph datasets with 10 participating clients.}
\label{curve}
\end{figure}

\textbf{Result 1: the answer to Q1}. The comparison results are presented in Table \ref{baseline_comp}. As observed, the proposed FedTAD consistently outperforms baselines. Specifically, compared with FedAvg, FedTAD brings at most 5.3\% performance improvement; Compared with FedGTA, which is the optimal baseline in most cases, FedTAD can achieve at most 4.9\% performance improvement. The convergence curves of FedTAD and baselines are shown in Fig.~\ref{curve}. As observed, FedTAD consistently converges within fewer communication rounds, demonstrating that FedTAD is suitable for subgraph-FL scenarios with limited communication overhead.

Moreover, to reduce privacy leaks, we also evaluated the performance of FedTAD when 10\% Gaussian noise is introduced into the uploaded class-level knowledge reliability. As observed, FedTAD exhibits a performance decline of less than 0.4\% and remains superior to the baseline.

\textbf{Result 2: the answer to Q2}. To answer \textbf{Q2}, we focus on the composition of FedTAD, including the class-wise knowledge reliability module on the client side, and the topology-aware knowledge distillation module on the server side. 

\begin{figure}
\rmfamily
\captionsetup{skip=0pt}
\centering
\subfigure[CKR Ablation]{
\includegraphics[width=0.23\textwidth]{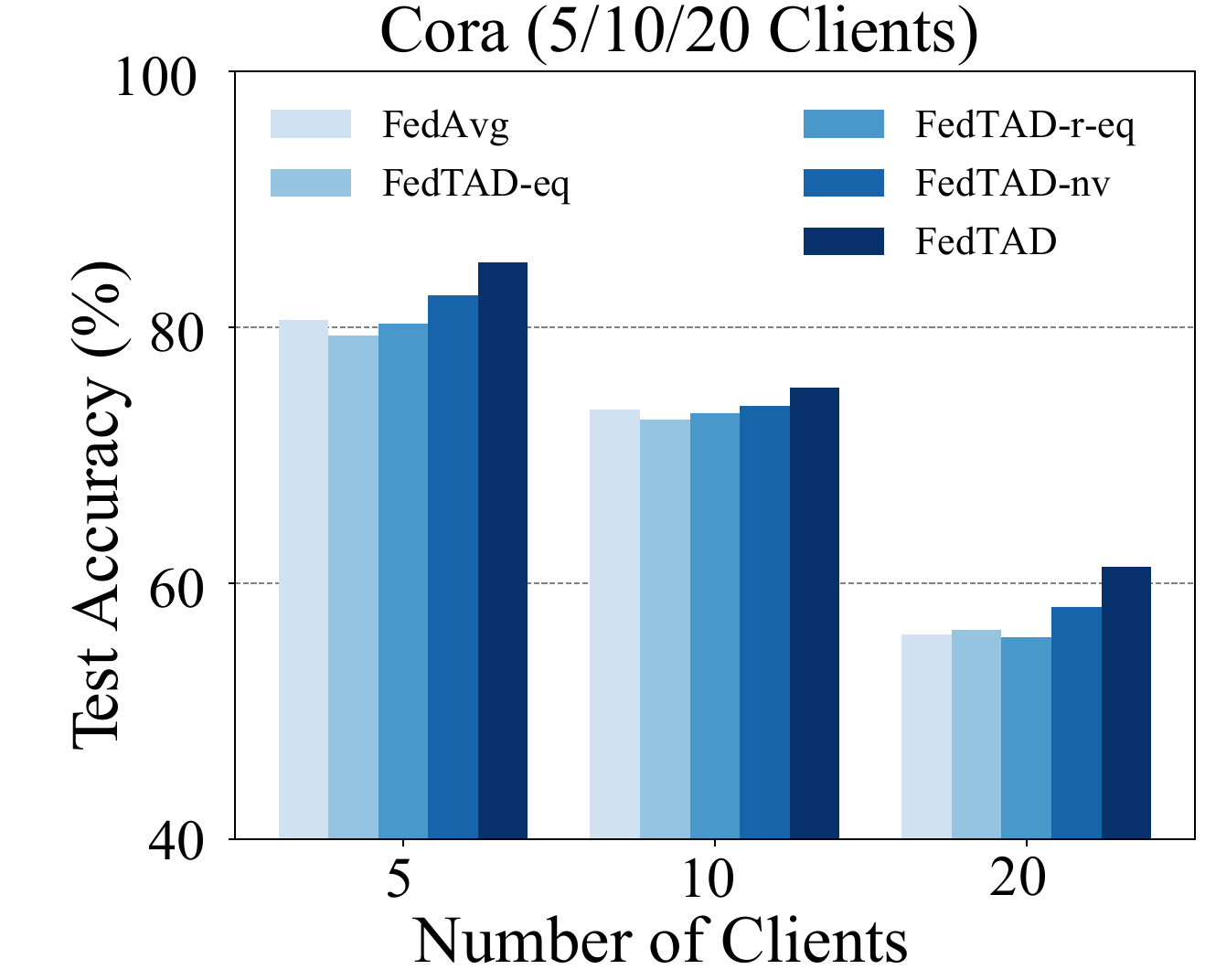} 
}\hspace{-2.3mm}
\subfigure[TAD Ablation]{
\includegraphics[width=0.23\textwidth]{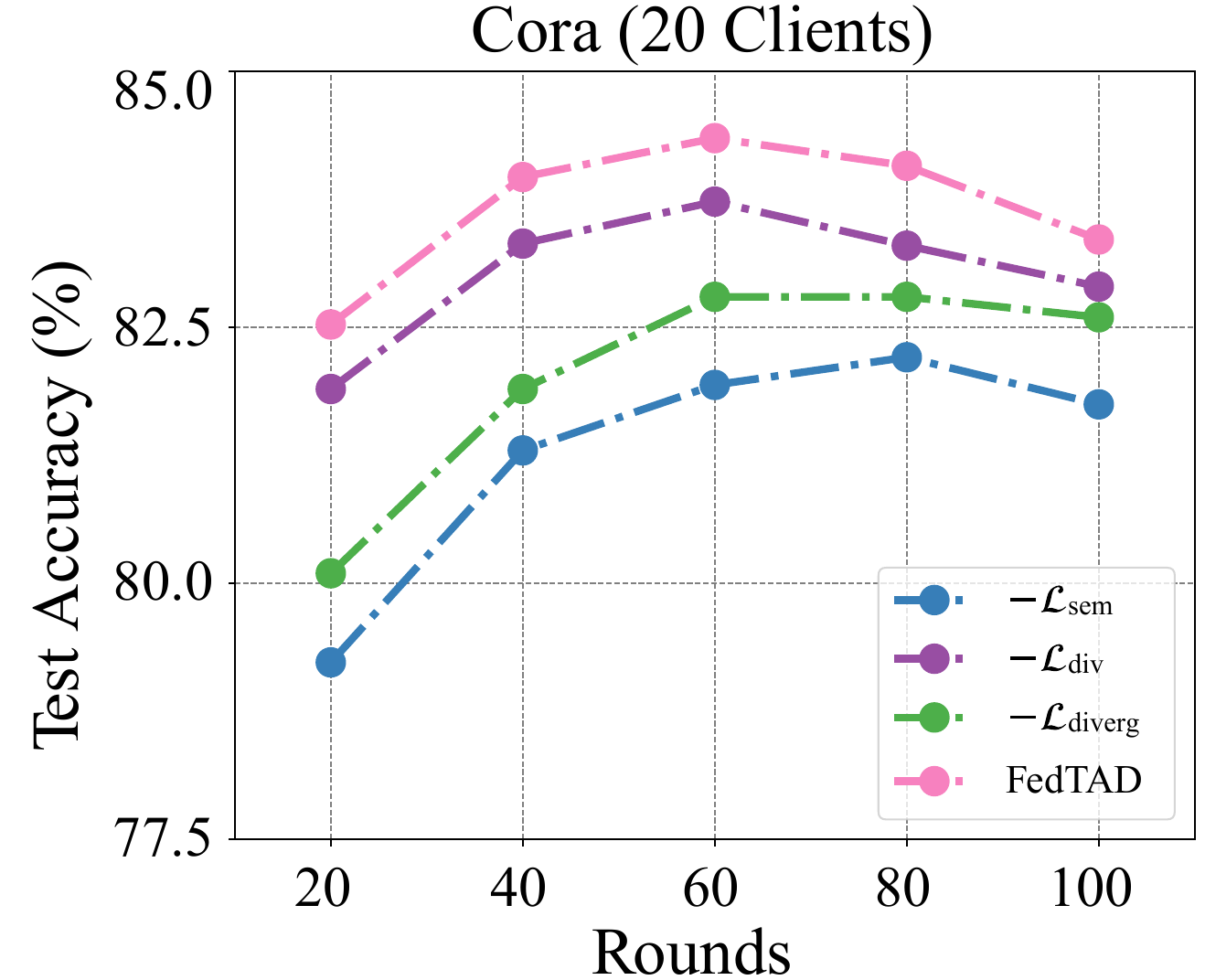}
}
\DeclareGraphicsExtensions.
\caption{Experimental results for the ablation study.}
\label{abl}
\end{figure}

First, we conducted an ablation study on the Cora datasets with 5/10/20 participating clients to demonstrate the effectiveness of class-level knowledge reliability (Eq.~\ref{class_wise_kr_eq}). Specifically, we performed two replacements for its calculation. (i) $\boldsymbol{\phi}_c = 1$ (each class is \textbf{eq}ually reliable, \textbf{FedTAD-eq} for short); (ii) $\boldsymbol{\phi}_c = |\mathcal{V}^{\text{labeled}}|$ (only considering \textbf{n}ode \textbf{v}ariation while ignoring topology variation, \textbf{FedTAD-nv} for short). Moreover, we additionally replace the pseudo graph in FedTAD-eq with the real global graph (\textbf{FedTAD-r-eq} for short). The comparative results are presented in Fig.~\ref{abl} (a). As observed, the performance achieved by FedTAD-eq, FedTAD-r-eq, and FedAvg are nearly identical, demonstrating that blindly conducting knowledge distillation without considering the class-wise knowledge reliability cannot improve the global model, even when relying on real graph data. Moreover, The performance of FedTAD-nv is consistently lower than FedTAD, suggesting that ignoring topology will lead to suboptimal class-wise knowledge reliability.

Afterward, we conducted an ablation study on the Cora dataset with 5 participating clients to investigate the contribution of each loss in the knowledge distillation module. Specifically, we remove three loss functions ($\mathcal{L}_{sem}$, $\mathcal{L}_{div}$ and $\mathcal{L}_{diverg}$) for the generator. The experimental results are shown in Fig.~\ref{abl} (b). As observed, removing any of the losses results in performance degradation for FedTAD, indicating that each loss function plays a crucial role in FedTAD. 

\begin{figure}
\rmfamily
\captionsetup{skip=0pt}
\centering
\subfigure[Cora (20 Clients)]{
\includegraphics[width=0.2\textwidth]{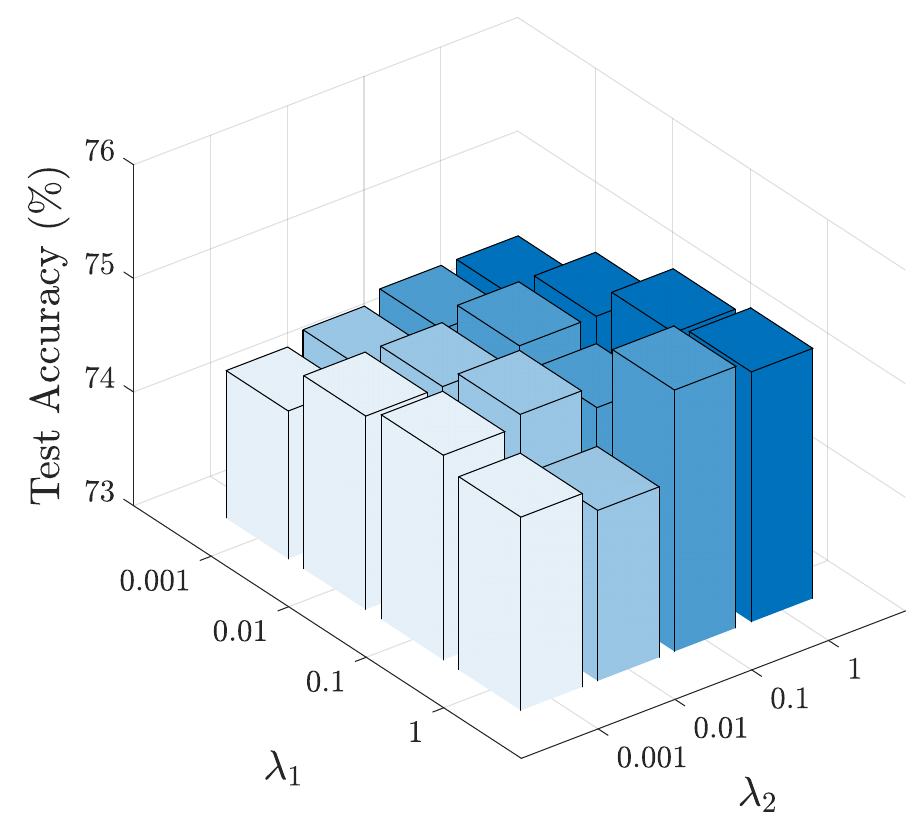} 
}
\subfigure[ogbn-arxiv (20 Clients)]{
\includegraphics[width=0.20\textwidth]{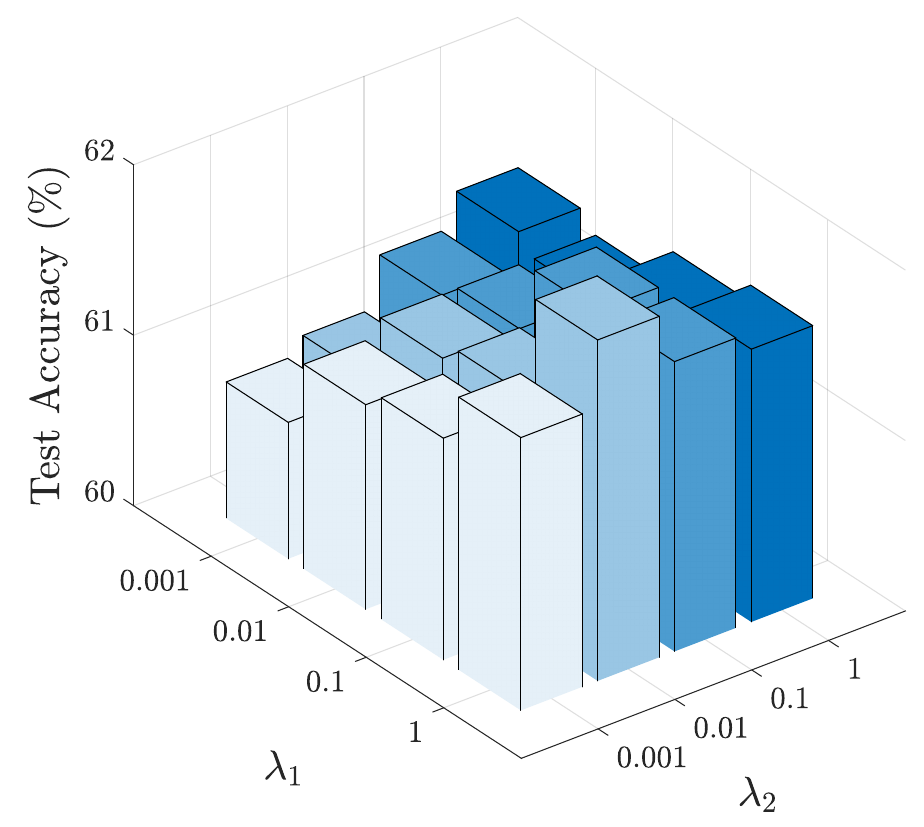}
}
\DeclareGraphicsExtensions.
\caption{Sensitive analysis for two trade-off parameters $\lambda_1$ and $\lambda_2$.}
\label{sen}
\end{figure}

\textbf{Result 3: the answer to Q3}. To answer \textbf{Q3}, we assess the performance of FedTAD with 20 participating clients under diverse combinations of trade-off parameters.
Specifically, we tune the trade-off parameters $\lambda_1$ and $\lambda_2$ within $\{10^{-3}, 10^{-2}, 10^{-1}, 1\}$. The sensitive analysis results are presented in Fig.~\ref{sen}. As observed, the performance fluctuation of FedTAD remains consistently below 1\%, demonstrating that FedTAD is insensitive to the value of trade-off parameters.

\textbf{Result 4: the answer to Q4}. To answer \textbf{Q4},  we compare FedTAD with four baseline methods, including FedAvg, Fed-PUB, FedSage+, and FedGTA on Cora and ogbn-arxiv datasets with 20 participating clients. The activate fraction varies in $\{0.2, 0.4, 0.6, 0.8, 1.0\}$. The experimental results are presented in Table.~\ref{tab_frac}. As observed, FedTAD consistently outperforms the baseline, indicating that FedTAD can maintain robustness with partial client participation.

\textbf{Result 5: the answer to Q5}. To answer \textbf{Q5}, we incorporate FedTAD as a plugin into several FGL (i.e., graph-level FL and subgraph-FL) baseline methods, including GCFL+, Fed-PUB, FedSage+, and FedGTA, and evaluate their performance boost on the ogbn-arxiv dataset with 20 participating clients. The experimental results are presented in Table.~\ref{tab_plug}. As observed, FedTAD consistently improves baselines and achieves at most a 5.1\% performance boost. 

\textbf{Result 6: the answer to Q6.} To answer \textbf{Q6}, we provide the complexity analysis of FedTAD. On the \underline{Client side}, calculating the topology-aware node embedding (Eq.~\ref{topo_emb_eq}) costs $\mathcal{O}(pB)$ with sparse computation, where $p$ denotes the walk distance, and $B$ denotes the number of generated nodes; class-wise knowledge reliability (Eq.~\ref{class_wise_kr_eq}) costs $\mathcal{O}(m)$, where $m$ denotes the number of edges; On the \underline{Server side}, the time complexity of model aggregation depends on the chosen optimization strategy; the topology-aware data-free knowledge distillation process costs $\mathcal{O}(I(I_gBf(z+B+Nc) + I_dBfNc))$, where $N$ denotes the number of participating clients, and $z, f, c$ denote the dimension of sampled noise, node feature, and number of classes, respectively.

\section{Related Work}
\paragraph{Graph Neural Networks.} 
    Earlier research on deep graph learning extends convolution to handle graphs~\cite{bruna2013spectral} but comes with notable parameter counts. To this end, GCN~\cite{kipf2016semi} simplifies graph convolution by utilizing a 1-order Chebyshev filter to capture local neighborhood information. Moreover, GAT~\cite{velivckovic2017graph} adopts graph attention, allowing weighted aggregation. GraphSAGE~\cite{hamilton2017inductive} introduces a variety of learnable aggregation functions for performing message aggregation. 
    Further details on GNN research can be found in surveys ~\cite{wu2020comprehensive,zhou2020graph}.

\paragraph{Federated Graph Learning.}
    Mainstream FGL studies can be divided into two settings: 
    (i) Graph-FL: GCFL+~\cite{xie2021federated} and FedStar~\cite{tan2023federated}.
    Each client collects multiple independent graphs, aiming to collaboratively solve graph-level downstream tasks (e.g., graph classification). 
    (ii) Subgraph-FL: each client holds a subgraph of an implicit global graph, aiming to solve node-level downstream tasks (e.g., node classification). Notably, there are currently two main challenges: 
    (a) The global GNN performance degradation caused by \textit{subgraph heterogeneity} (node and topology variation in multi-client subgraphs).
    Fed-PUB \cite{baek2023personalized} points out this heterogeneity stems from the various label distributions and addresses it through personalized technologies. FedGTA \cite{xkLi_FedGTA_VLDB_2024} utilizes a graph mixed moments to identify similar subgraphs, thereby achieving topology-aware aggregation. (b) The information loss caused by \textit{missing edges} (connections implicitly exist between subgraphs across clients are lost due to distributed storage). 
    Approaches for handling missing edges include FedSage+ \cite{zhang2021subgraph} and FedGNN~\cite{wu2021fedgnn}.

\begin{table}
\rmfamily
\centering
\caption{\textrm{Performance boost on baselines with FedTAD plug.}}
\label{tab_plug}
\resizebox{0.48\textwidth}{!}{
\begin{tabular}{c|ccccc} 
\toprule
Dataset $(\rightarrow)$&   \multicolumn{2}{c}{Cora (20 Clients)} & &\multicolumn{2}{c}{ogbn-arxiv (20 Clients)}\\
\cline{1-1} \cline{2-3} \cline{5-6}
\rule{0pt}{1.1em}Method $(\downarrow)$     & w/o FedTAD &  with FedTAD && w/o FedTAD &  with FedTAD\\
\midrule
GCFL+       &  56.6±0.2 & 61.5±0.1 & &   55.9±0.4 & 59.4±0.2  \\
Fed-PUB  &   56.5±0.5 &  61.6±0.2 &&   56.1±0.1 &  59.7±0.7 \\
FedSage+ &   58.1±0.7 &  61.7±0.4 &&   56.3±0.4 &   60.1±0.3\\
FedGTA     &   58.8±0.2 &  62.5±0.2 &&   57.3±0.1 &  60.5±0.4\\
\bottomrule
\end{tabular}
}
\end{table}
\begin{table}
\rmfamily
\centering
\caption{\textrm{Performance of FedTAD on various active fractions.}}

\label{tab_frac}
\resizebox{0.48\textwidth}{!}{
\begin{tabular}{c|ccccc} 
\toprule
Dataset $(\rightarrow)$&   \multicolumn{5}{c}{Cora (20 Clients)}\\
\cline{1-1} \cline{2-6}
\rule{0pt}{1.1em}Method $(\downarrow)$     &  20\% frac. &  40\% frac. & 60\% frac. & 80\% frac. & 100\% frac.\\
\midrule
FedAvg       &   44.5±0.4 &  50.3±0.2 & 51.8±0.3 & 53.1±0.6 & 56.0±0.3 \\
Fed-PUB &   46.1±0.6 &  51.9±0.3 & 52.1±0.5 & 53.2±0.3 & 56.5±0.5 \\
FedSage+ &   \cellcolor{gray!20}47.5±0.5 &  \cellcolor{gray!20}52.8±0.2 & \cellcolor{gray!20}54.4±0.2 & \underline{56.4±0.3} & \cellcolor{gray!20}58.1±0.7 \\
FedGTA     & \underline{49.6±0.3}  &  \underline{53.1±0.3} & \underline{55.7±0.2} & \cellcolor{gray!20}56.2±0.4 & \underline{58.8±0.2} \\
FedTAD      &  \textbf{51.5±0.2} & \textbf{54.2±0.4} & \textbf{57.4±0.4} & \textbf{58.5±0.4} & \textbf{61.3±0.3} \\
\bottomrule
\end{tabular}
}
\vspace{-8pt}
\end{table}

\paragraph{Data-Free Knowledge Distillation.}
    This is a knowledge distillation technique that transfers the knowledge of the pre-trained teacher model to the student model through generated pseudo data instead of accessing the original data, including DeepImpression \cite{nayak2019zero}, DeepInversion \cite{yin2020dreaming}, and DFAL \cite{chen2019data}. 
    Recently, some studies introduced this strategy into FL to alleviate data heterogeneity, such as FedGen \cite{zhu2021data} and FedFTG \cite{zhang2022fine}. Unfortunately, they cannot be adapted to subgraph-FL scenarios as they can only generate Euclidean data rather than graph data.

\section{Conclusion}

In this paper, we investigate the subgraph heterogeneity problem in subgraph-FL. To the best of our knowledge, we are the first to provide an in-depth investigation of the subgraph heterogeneity problem by individually analyzing the impact of node and topology variation. We demonstrate that subgraph heterogeneity results in variation in the class-wise knowledge reliability. Building on this insight, we propose FedTAD, utilizing topology-aware data-free knowledge distillation to enable the local models to transfer their most reliable knowledge to correct the global model misled by unreliable knowledge. The experimental results demonstrate that FedTAD significantly outperforms state-of-the-art baselines.

\newpage
\section*{Acknowledgments}
This work was supported by the Shenzhen Science and Technology Program (Grant No.~KJZD20230923113901004), the Science and Technology Planning Project of Guangdong Province under Grant 2023A0505020006, and the National Natural Science Foundation of China under Grants U2001209, 62072486.

\bibliographystyle{named}
\bibliography{ijcai23}

\begin{thebibliography}{}

\bibitem[\protect\citeauthoryear{Akiba \bgroup \em et al.\egroup
  }{2019}]{akiba2019optuna}
Takuya Akiba, Shotaro Sano, Toshihiko Yanase, Takeru Ohta, and Masanori Koyama.
\newblock Optuna: A next-generation hyperparameter optimization framework.
\newblock In {\em Proceedings of the 25th ACM SIGKDD international conference
  on knowledge discovery \& data mining}, pages 2623--2631, 2019.

\bibitem[\protect\citeauthoryear{Baek \bgroup \em et al.\egroup
  }{2023}]{baek2023personalized}
Jinheon Baek, Wonyong Jeong, Jiongdao Jin, Jaehong Yoon, and Sung~Ju Hwang.
\newblock Personalized subgraph federated learning.
\newblock In {\em International Conference on Machine Learning}, pages
  1396--1415. PMLR, 2023.

\bibitem[\protect\citeauthoryear{Blondel \bgroup \em et al.\egroup
  }{2008}]{blondel2008fast}
Vincent~D Blondel, Jean-Loup Guillaume, Renaud Lambiotte, and Etienne Lefebvre.
\newblock Fast unfolding of communities in large networks.
\newblock {\em Journal of statistical mechanics: theory and experiment},
  2008(10):P10008, 2008.

\bibitem[\protect\citeauthoryear{Bruna \bgroup \em et al.\egroup
  }{2013}]{bruna2013spectral}
Joan Bruna, Wojciech Zaremba, Arthur Szlam, and Yann LeCun.
\newblock Spectral networks and locally connected networks on graphs.
\newblock {\em arXiv preprint arXiv:1312.6203}, 2013.

\bibitem[\protect\citeauthoryear{Chen \bgroup \em et al.\egroup
  }{2019}]{chen2019data}
Hanting Chen, Yunhe Wang, Chang Xu, Zhaohui Yang, Chuanjian Liu, Boxin Shi,
  Chunjing Xu, Chao Xu, and Qi~Tian.
\newblock Data-free learning of student networks.
\newblock In {\em Proceedings of the IEEE/CVF international conference on
  computer vision}, pages 3514--3522, 2019.

\bibitem[\protect\citeauthoryear{Ding \bgroup \em et al.\egroup
  }{2018}]{ding2018semi}
Ming Ding, Jie Tang, and Jie Zhang.
\newblock Semi-supervised learning on graphs with generative adversarial nets.
\newblock In {\em Proceedings of the 27th ACM International Conference on
  Information and Knowledge Management}, pages 913--922, 2018.

\bibitem[\protect\citeauthoryear{Fu \bgroup \em et al.\egroup
  }{2022}]{fu2022federated}
Xingbo Fu, Binchi Zhang, Yushun Dong, Chen Chen, and Jundong Li.
\newblock Federated graph machine learning: A survey of concepts, techniques,
  and applications.
\newblock {\em ACM SIGKDD Explorations Newsletter}, 24(2):32--47, 2022.

\bibitem[\protect\citeauthoryear{Gao \bgroup \em et al.\egroup
  }{2022}]{gao2022feddc}
Liang Gao, Huazhu Fu, Li~Li, Yingwen Chen, Ming Xu, and Cheng-Zhong Xu.
\newblock Feddc: Federated learning with non-iid data via local drift
  decoupling and correction.
\newblock In {\em Proceedings of the IEEE/CVF conference on computer vision and
  pattern recognition}, pages 10112--10121, 2022.

\bibitem[\protect\citeauthoryear{Gilmer \bgroup \em et al.\egroup
  }{2017}]{gilmer2017neural}
Justin Gilmer, Samuel~S Schoenholz, Patrick~F Riley, Oriol Vinyals, and
  George~E Dahl.
\newblock Neural message passing for quantum chemistry.
\newblock In {\em International conference on machine learning}, pages
  1263--1272. PMLR, 2017.

\bibitem[\protect\citeauthoryear{Guo and Wang}{2020}]{guo2020deep}
Zhiwei Guo and Heng Wang.
\newblock A deep graph neural network-based mechanism for social
  recommendations.
\newblock {\em IEEE Transactions on Industrial Informatics}, 17(4):2776--2783,
  2020.

\bibitem[\protect\citeauthoryear{Hamilton \bgroup \em et al.\egroup
  }{2017}]{hamilton2017inductive}
Will Hamilton, Zhitao Ying, and Jure Leskovec.
\newblock Inductive representation learning on large graphs.
\newblock {\em Advances in neural information processing systems}, 30, 2017.

\bibitem[\protect\citeauthoryear{He \bgroup \em et al.\egroup
  }{2021}]{he2021fedgraphnn}
Chaoyang He, Keshav Balasubramanian, Emir Ceyani, Carl Yang, Han Xie, Lichao
  Sun, Lifang He, Liangwei Yang, Philip~S Yu, Yu~Rong, et~al.
\newblock Fedgraphnn: A federated learning system and benchmark for graph
  neural networks.
\newblock {\em arXiv preprint arXiv:2104.07145}, 2021.

\bibitem[\protect\citeauthoryear{Hu \bgroup \em et al.\egroup
  }{2020}]{hu2020open}
Weihua Hu, Matthias Fey, Marinka Zitnik, Yuxiao Dong, Hongyu Ren, Bowen Liu,
  Michele Catasta, and Jure Leskovec.
\newblock Open graph benchmark: Datasets for machine learning on graphs.
\newblock {\em Advances in neural information processing systems},
  33:22118--22133, 2020.

\bibitem[\protect\citeauthoryear{Karimireddy \bgroup \em et al.\egroup
  }{2020}]{karimireddy2020scaffold}
Sai~Praneeth Karimireddy, Satyen Kale, Mehryar Mohri, Sashank Reddi, Sebastian
  Stich, and Ananda~Theertha Suresh.
\newblock Scaffold: Stochastic controlled averaging for federated learning.
\newblock In {\em International conference on machine learning}, pages
  5132--5143. PMLR, 2020.

\bibitem[\protect\citeauthoryear{Kipf and Welling}{2016}]{kipf2016semi}
Thomas~N Kipf and Max Welling.
\newblock Semi-supervised classification with graph convolutional networks.
\newblock {\em arXiv preprint arXiv:1609.02907}, 2016.

\bibitem[\protect\citeauthoryear{Li \bgroup \em et al.\egroup
  }{2020}]{li2020federated}
Tian Li, Anit~Kumar Sahu, Manzil Zaheer, Maziar Sanjabi, Ameet Talwalkar, and
  Virginia Smith.
\newblock Federated optimization in heterogeneous networks.
\newblock {\em Proceedings of Machine learning and systems}, 2:429--450, 2020.

\bibitem[\protect\citeauthoryear{Li \bgroup \em et al.\egroup
  }{2021}]{li2021model}
Qinbin Li, Bingsheng He, and Dawn Song.
\newblock Model-contrastive federated learning.
\newblock In {\em Proceedings of the IEEE/CVF conference on computer vision and
  pattern recognition}, pages 10713--10722, 2021.

\bibitem[\protect\citeauthoryear{Li \bgroup \em et al.\egroup
  }{2023}]{xkLi_FedGTA_VLDB_2024}
Xunkai Li, Zhengyu Wu, Wentao Zhang, Yinlin Zhu, Ronghua Li, and Guoren Wang.
\newblock Fedgta: Topology-aware averaging for federated graph learning.
\newblock {\em Proc. {VLDB} Endow.}, 17(1):41--50, 2023.

\bibitem[\protect\citeauthoryear{Mao \bgroup \em et al.\egroup
  }{2019}]{mao2019mode}
Qi~Mao, Hsin-Ying Lee, Hung-Yu Tseng, Siwei Ma, and Ming-Hsuan Yang.
\newblock Mode seeking generative adversarial networks for diverse image
  synthesis.
\newblock In {\em Proceedings of the IEEE/CVF conference on computer vision and
  pattern recognition}, pages 1429--1437, 2019.

\bibitem[\protect\citeauthoryear{McMahan \bgroup \em et al.\egroup
  }{2017}]{mcmahan2017communication}
Brendan McMahan, Eider Moore, Daniel Ramage, Seth Hampson, and Blaise~Aguera
  y~Arcas.
\newblock Communication-efficient learning of deep networks from decentralized
  data.
\newblock In {\em Artificial intelligence and statistics}, pages 1273--1282.
  PMLR, 2017.

\bibitem[\protect\citeauthoryear{Nayak \bgroup \em et al.\egroup
  }{2019}]{nayak2019zero}
Gaurav~Kumar Nayak, Konda~Reddy Mopuri, Vaisakh Shaj, Venkatesh~Babu
  Radhakrishnan, and Anirban Chakraborty.
\newblock Zero-shot knowledge distillation in deep networks.
\newblock In {\em International Conference on Machine Learning}, pages
  4743--4751. PMLR, 2019.

\bibitem[\protect\citeauthoryear{Shchur \bgroup \em et al.\egroup
  }{2018}]{shchur2018pitfalls}
Oleksandr Shchur, Maximilian Mumme, Aleksandar Bojchevski, and Stephan
  G{\"u}nnemann.
\newblock Pitfalls of graph neural network evaluation.
\newblock {\em arXiv preprint arXiv:1811.05868}, 2018.

\bibitem[\protect\citeauthoryear{Sun \bgroup \em et al.\egroup
  }{2023}]{sun2023breaking}
Henan Sun, Xunkai Li, Zhengyu Wu, Daohan Su, Rong-Hua Li, and Guoren Wang.
\newblock Breaking the entanglement of homophily and heterophily in
  semi-supervised node classification, 2023.

\bibitem[\protect\citeauthoryear{Tan \bgroup \em et al.\egroup
  }{2023}]{tan2023federated}
Yue Tan, Yixin Liu, Guodong Long, Jing Jiang, Qinghua Lu, and Chengqi Zhang.
\newblock Federated learning on non-iid graphs via structural knowledge
  sharing.
\newblock In {\em Proceedings of the AAAI conference on artificial
  intelligence}, volume~37, pages 9953--9961, 2023.

\bibitem[\protect\citeauthoryear{Tong \bgroup \em et al.\egroup
  }{2006}]{tong2006fast}
Hanghang Tong, Christos Faloutsos, and Jia-Yu Pan.
\newblock Fast random walk with restart and its applications.
\newblock In {\em Sixth international conference on data mining (ICDM'06)},
  pages 613--622. IEEE, 2006.

\bibitem[\protect\citeauthoryear{Veli{\v{c}}kovi{\'c} \bgroup \em et al.\egroup
  }{2017}]{velivckovic2017graph}
Petar Veli{\v{c}}kovi{\'c}, Guillem Cucurull, Arantxa Casanova, Adriana Romero,
  Pietro Lio, and Yoshua Bengio.
\newblock Graph attention networks.
\newblock {\em arXiv preprint arXiv:1710.10903}, 2017.

\bibitem[\protect\citeauthoryear{Wu \bgroup \em et al.\egroup
  }{2020}]{wu2020comprehensive}
Zonghan Wu, Shirui Pan, Fengwen Chen, Guodong Long, Chengqi Zhang, and S~Yu
  Philip.
\newblock A comprehensive survey on graph neural networks.
\newblock {\em IEEE transactions on neural networks and learning systems},
  32(1):4--24, 2020.

\bibitem[\protect\citeauthoryear{Wu \bgroup \em et al.\egroup
  }{2021}]{wu2021fedgnn}
Chuhan Wu, Fangzhao Wu, Yang Cao, Yongfeng Huang, and Xing Xie.
\newblock Fedgnn: Federated graph neural network for privacy-preserving
  recommendation.
\newblock {\em arXiv preprint arXiv:2102.04925}, 2021.

\bibitem[\protect\citeauthoryear{Xie \bgroup \em et al.\egroup
  }{2021}]{xie2021federated}
Han Xie, Jing Ma, Li~Xiong, and Carl Yang.
\newblock Federated graph classification over non-iid graphs.
\newblock {\em Advances in Neural Information Processing Systems},
  34:18839--18852, 2021.

\bibitem[\protect\citeauthoryear{Yang \bgroup \em et al.\egroup
  }{2016}]{yang2016revisiting}
Zhilin Yang, William Cohen, and Ruslan Salakhudinov.
\newblock Revisiting semi-supervised learning with graph embeddings.
\newblock In {\em International conference on machine learning}, pages 40--48.
  PMLR, 2016.

\bibitem[\protect\citeauthoryear{Yang \bgroup \em et al.\egroup
  }{2021}]{yang2021financial}
Shuo Yang, Zhiqiang Zhang, Jun Zhou, Yang Wang, Wang Sun, Xingyu Zhong, Yanming
  Fang, Quan Yu, and Yuan Qi.
\newblock Financial risk analysis for smes with graph-based supply chain
  mining.
\newblock In {\em Proceedings of the Twenty-Ninth International Conference on
  International Joint Conferences on Artificial Intelligence}, pages
  4661--4667, 2021.

\bibitem[\protect\citeauthoryear{Yin \bgroup \em et al.\egroup
  }{2020}]{yin2020dreaming}
Hongxu Yin, Pavlo Molchanov, Jose~M Alvarez, Zhizhong Li, Arun Mallya, Derek
  Hoiem, Niraj~K Jha, and Jan Kautz.
\newblock Dreaming to distill: Data-free knowledge transfer via deepinversion.
\newblock In {\em Proceedings of the IEEE/CVF Conference on Computer Vision and
  Pattern Recognition}, pages 8715--8724, 2020.

\bibitem[\protect\citeauthoryear{Zhang \bgroup \em et al.\egroup
  }{2021}]{zhang2021subgraph}
Ke~Zhang, Carl Yang, Xiaoxiao Li, Lichao Sun, and Siu~Ming Yiu.
\newblock Subgraph federated learning with missing neighbor generation.
\newblock {\em Advances in Neural Information Processing Systems},
  34:6671--6682, 2021.

\bibitem[\protect\citeauthoryear{Zhang \bgroup \em et al.\egroup
  }{2022}]{zhang2022fine}
Lin Zhang, Li~Shen, Liang Ding, Dacheng Tao, and Ling-Yu Duan.
\newblock Fine-tuning global model via data-free knowledge distillation for
  non-iid federated learning.
\newblock In {\em Proceedings of the IEEE/CVF conference on computer vision and
  pattern recognition}, pages 10174--10183, 2022.

\bibitem[\protect\citeauthoryear{Zhou \bgroup \em et al.\egroup
  }{2020}]{zhou2020graph}
Jie Zhou, Ganqu Cui, Shengding Hu, Zhengyan Zhang, Cheng Yang, Zhiyuan Liu,
  Lifeng Wang, Changcheng Li, and Maosong Sun.
\newblock Graph neural networks: A review of methods and applications.
\newblock {\em AI Open}, 1:57--81, 2020.

\bibitem[\protect\citeauthoryear{Zhu \bgroup \em et al.\egroup
  }{2020}]{zhu2020beyond}
Jiong Zhu, Yujun Yan, Lingxiao Zhao, Mark Heimann, Leman Akoglu, and Danai
  Koutra.
\newblock Beyond homophily in graph neural networks: Current limitations and
  effective designs.
\newblock {\em Advances in neural information processing systems},
  33:7793--7804, 2020.

\bibitem[\protect\citeauthoryear{Zhu \bgroup \em et al.\egroup
  }{2021}]{zhu2021data}
Zhuangdi Zhu, Junyuan Hong, and Jiayu Zhou.
\newblock Data-free knowledge distillation for heterogeneous federated
  learning.
\newblock In {\em International conference on machine learning}, pages
  12878--12889. PMLR, 2021.

\end{thebibliography}

\end{document}